\documentclass[x11names,dvipsnames]{neus2025}

\title{Pretrained Embeddings as a Behavior Specification Mechanism}
\usepackage{times}
\usepackage{graphicx,verbatimbox}
\usepackage{subcaption}
\usepackage{algorithmic}
\usepackage{algorithm}
\usepackage{array}
\usepackage{enumitem}
\usepackage[labelfont=bf,margin=0.25in,font=small]{caption}
\usepackage{textcomp}
\usepackage{stfloats}
\usepackage{url}
\usepackage{verbatim}
\usepackage{graphicx}
\usepackage{nicematrix}
\usepackage{tikz}
\usepackage{color}
\usepackage{booktabs}
\usepackage{arydshln}
\usepackage{array}

\DeclareMathOperator*{\argmin}{arg\,min}

\setlist[itemize]{leftmargin=3mm}

\newcommand{\eventually}[0]{\textbf{F}}
\newcommand{\always}[0]{\textbf{G}}
\newcommand{\embeddings}[0]{\mathcal{Z}}
\newcommand{\metrics}[0]{\mathbb{M}}
\newcommand{\until}[0]{\textbf{U}}
\newcommand{\trace}[0]{\sigma}
\newcommand{\rob}[0]{\rho}
\newcommand{\bound}[0]{\textbf{b}}

\newcolumntype{C}{>{\centering\arraybackslash}p{3.5em}}

\author{%
 \Name{Parv Kapoor} \Email{parvk@andrew.cmu.edu}\\
 \addr Carnegie Mellon University
 \AND
 \Name{Abigail Hammer} \Email{arhammer@andrew.cmu.edu}\\
 \addr Carnegie Mellon University
 \AND
 \Name{Ashish Kapoor} \Email{ashish@scaledfoundations.ai}\\
 \addr Scaled Foundations
 \AND
 \Name{Karen Leung} \Email{kymleung@uw.edu}\\
 \addr University of Washington
 \AND
 \Name{Eunsuk Kang} \Email{eunsukk@andrew.cmu.edu}\\
 \addr Carnegie Mellon University
}

\begin{document}

\maketitle

\begin{abstract}
We propose an approach to formally specify the behavioral properties of systems that rely on a perception model for interactions with the physical world. The key idea is to introduce \emph{embeddings}---mathematical representations of a real-world concept---as a first-class construct in a  specification language, where properties are expressed in terms of distances between a pair of ideal and observed embeddings. To realize this approach, we propose a new type of temporal logic called \emph{Embedding Temporal Logic (ETL)}, and describe how it can be used to express a wider range of properties about AI-enabled systems than previously possible. We demonstrate the applicability of ETL through a preliminary evaluation involving planning tasks in robots that are driven by foundation models; the results are promising, showing that embedding-based specifications can be used to steer a system towards desirable behaviors.
\end{abstract}

\section{Introduction}

Modern artificial intelligence (AI) technologies, such as foundation models (FMs), are rapidly merging as a key component in autonomous systems, being used to perform critical functions such as perception and planning. Techniques for achieving high assurance, such as verification and run-time monitoring, rely on the availability of \emph{formal specifications} that capture the desired properties of a system. However, formally specifying the behavior of an AI-enabled system remains an open challenge~\cite{seshia-atva18}. 

Formal specifications, especially those written in a temporal logic, are expressed in terms of propositions about parts of the system state that can be observed or estimated  through sensors (e.g., the velocity of a robot). For autonomous systems, behavioral properties often refer to interactions with physical objects (e.g., ``the robot should avoid colliding with a table''); to observe these objects, the system relies on a perception model that operates over a high-dimensional input image. A formal specification for such a system would require propositions that relate physical objects to the input image. Here lies the fundamental obstacle to specification: devising a precise, mathematical encoding of a physical object (e.g., a table) over the high-dimensional input space (e.g., pixels) is likely to be difficult, if not impossible.

In this paper, we propose a new approach for formally specifying the behavioral properties of an AI-based system. The key idea is to introduce \emph{embeddings}---mathematical  representations of real-world objects---as a first-class concept in a specification notation, and express a property in terms of \emph{distances} between a \emph{target} embedding (an ideal representation of the world for the system to reach or avoid, given as part of the specification) and an \emph{observed} embedding (a representation observed and generated through a sensor during system execution). For example, consider a rescue robot that is tasked with satisfying the following requirement: ``Locate a potential victim while avoiding areas with fire.'' Such a property would be difficult to specify using an existing specification language, due to its dependence on  perception; even with access to accurate localization, the exact locations of these real-world objects are often unknown and dynamic. Instead, in an embedding-based approach, this task may be expressed as ``reach the state of the world in which the system observes an object that closely resembles a potential victim, while avoiding those states where the observation resembles an area with fire.'' Such a specification could then be used, for example, as part of a run-time monitor or a planner to ensure that the system conforms to the desired property.

As a realization of this approach, we propose \emph{Embedding Temporal Logic (ETL)}, a temporal logic for specifying the behaviors of AI-based systems (Section~\ref{sec:etl}). Compared to state-based temporal specifications (such as linear temporal logic (LTL)~\cite{pnueli77}), which are evaluated over a sequence of states, an ETL specification is evaluated over a \emph{sequence of embeddings}, where each embedding is created from an observation that the system makes at a particular point in its execution. As counterparts to propositions in LTL, atomic constructs in ETL are \emph{embedding predicates}, which impose a constraint over the distance between a pair of embeddings; e.g., $dist(z_o, z_t) \leq \epsilon$, where $z_t$ and $z_o$ are the target and observed embeddings, respectively, and $\epsilon \in \mathbb{M}$ is a given \emph{distance threshold} in  metric space $\metrics{}$. A target embedding in a predicate is specified by providing  images or text that correspond to a real-world concept (e.g., images or a textual description of fire). Standard temporal operators (e.g., $\always, \eventually$) are used to construct temporal properties out of atomic predicates.

We believe that \textbf{embeddings as a specification mechanism} have potential to \textbf{significantly broaden the range of properties that can be specified using a formal specification language}, facilitate the development of new assurance methods, and enable existing methods to be applied to AI-enabled systems. Additionally, leveraging embeddings for specification allows us to use the power of large, pretrained FMs \cite{radford2021learningtransferablevisualmodels, firoozi2023foundationmodelsroboticsapplications}—which excel at encoding high-level features and concepts into embedding spaces. Moreover, integrating world models \cite{DBLP:journals/corr/abs-1803-10122} that evolve within these embedding spaces further enriches our ability to reason about and verify the dynamic behaviors of such systems.
As a potential application, we demonstrate how ETL can be used for \emph{planning} tasks in robotic systems that use an FM for scene understanding and behavioral prediction (Section~\ref{sec:planning}). In particular, we propose a planning method that generates actions towards the goal of satisfying a given ETL specification. 
Our preliminary evaluation shows that the proposed method can effectively steer the system towards behaviors that are desirable with respect to the given specification (Section~\ref{sec:evaluation}).
 

%

\section{Background}

We briefly describe the ideas of pretrained vision models and world models, which play an important role for embeddings and FM-based planning, respectively. 

\label{sec:background}

\paragraph{Pretrained Vision Models.}
Pretrained vision models such as CLIP \cite{radford2021learning}, DinoV2 \cite{oquab2024dinov2learningrobustvisual} have revolutionized the field of computer vision by providing robust, high-level representations that capture rich semantic information. These models are generally trained on large-scale datasets and use supervised or self-supervised learning methods. The resulting representations (embeddings or feature vectors) have demonstrated impressive transferability across a wide range of downstream tasks including classification, detection, segmentation, and scene understanding. These models are built with convolutional neural networks (CNNs) \cite{NIPS2012_c399862d} or more recently vision transformers (ViTs) \cite{dosovitskiy2021imageworth16x16words}. ViT based models generally take as input a raw image and generate patch embeddings capturing granular information.

\paragraph{World Models.}
In the context of \emph{sequential decision-making} for robots---the process of generating a sequence of  actions to enable a system to perform a desired task---a robot must be able to predict future outcomes given their decisions.
Such predictive models are generally referred to as \textit{world models} \cite{DBLP:journals/corr/abs-1803-10122}, describing \textit{how the world evolves over time}. 
Abstractly, the ``world'' includes the robot itself, the surrounding environment, and interactions between the two.
However, in unstructured settings without accurate state estimation and scene understanding, developing a world model operating on observations (e.g., camera images) can be challenging.

Many recent works look to develop a world model by leveraging large amounts of offline observation data (e.g., images) collected from the system moving in the real-world and learning the transition dynamics of the system implicitly through a lower-dimensional embedding space \cite{DBLP:journals/corr/abs-1803-10122, watter2015embedcontrollocallylinear}. We follow a similar set up as described in \cite{zhou2025dinowm}.
The idea is to encode the high-dimensional observations during a time window $H$, (starting from time $t-H$ to $t$, denoted $o_{t-H:t}$),%
into corresponding embedded states $z_{t-H:t}$.
The embedded states and control sequence $a_{t-H:t}$ are inputs to a learned transition model $p$ to predict the latent state in the next time step $z_{t+1}$. A decoder $q$ can transform embedded states back into observation states $\hat{o}$---the decoder is not strictly needed in describing how the world evolves, but rather is more useful for visualizing what observation the latent state corresponds to. Mathematically, a world model consists of the following three components,
\begin{equation}
    \underbrace{z_t \sim \mathrm{enc}_\theta(z_t \mid o_t)}_{\text{Observation model}}, \quad \underbrace{z_{t+1} \sim p_\theta(z_{t+1} \mid z_{t-H:t}, a_{t-H:t})}_{\text{Transition model}}, \quad \underbrace{\hat{o}_t \sim q_\theta(o_t \mid z_t)}_{\text{Decoder model}},
\end{equation}

where $\theta$ denotes the parameters of the encoder (enc), decoder, and transition models. Essentially, at any time step $t$, we can compute the latent state $z_t$, and autoregressively roll out the transition model to simulate how the world evolves given a sequence of actions.

\section{Embedding Temporal Logic}
\label{sec:etl}

\subsection{Syntax and Semantics}
In our approach, a system is assumed to make an observation about the real world through a sensor (e.g., camera) at each step in its execution. The system also contains an \emph{encoder} that converts each observation into an embedding $z \in \mathcal{Z}$. Conceptually, $z$ is an approximation of a \emph{latent variable}; i.e., the state of the world that is only indirectly observable by the system. The underlying representation of embeddings (typically as a vector of a fixed size) depends on the encoder and other models that subsequently consume this embedding for further tasks, such as object identification and prediction. 

In ETL, a possible execution of the system, or a \emph{trace}, is represented as an infinite sequence $\trace$ of embeddings (i.e., $\trace = z_0, z_1,...$); we write $z_i$ to denote the embedding at the index $i$. We also assume the existence of \emph{distance function} $dist : \embeddings{} \times \embeddings{} \rightarrow \metrics$, which computes a distance between a pair of embeddings. The metric space depends on the embedding representation and the distance function used; a common option is $\mathbb{R}^{\geq 0}$. The syntax of ETL is as follows:
\begin{align*}
\varphi := u \;|\; \neg \varphi \;|\; \varphi_1 \land \varphi_2 \;|\; \varphi_1
  \until{} \varphi_2 
\end{align*}
where $u$ is a predicate of the form $f_u(z) > 0$ and $f_u(z)$ is the function (associated with   $u$) that maps the current embedding $z$ to a value in the metrics domain $\metrics{}$. As in LTL, the until operator
$\until{}$ can be used to express the \emph{eventually}
($\eventually{}$) and \emph{always} ($\always{}$):
$\eventually{} \varphi = \texttt{True} \until{} \varphi$
and
$\always{} \varphi = \neg \eventually{} \neg \varphi$. 

The semantics of ETL, defining the satisfaction of a formula by trace $\sigma$ at index $i$, also closely mirrors that of LTL:
\begin{align*}
    &\sigma, i \models u \iff f_u(z_i) > 0 \\
    &\sigma, i \models \neg \varphi \iff \sigma, i \not\models \varphi \\
    &\sigma, i \models \varphi_1 \land \varphi_2 \iff \sigma, i \models \varphi_1 \text{ and } \sigma, i \models \varphi_2 \\
    &\sigma, i \models \varphi_1 \until{} \varphi_2 \iff \exists j \geq i \text{ such that } \sigma, j \models \varphi_2 \text{ and } \forall k \text{ with } i \leq k < j, \sigma, k \models \varphi_1 
\end{align*}

\paragraph{Example 1: Reach and Avoid.} 
A common task in robotic systems involves reaching a particular goal (i.e., a location); e.g., "a system shall eventually navigate to the desired location" while avoiding unsafe regions. The unsafe region requirement can be described at a high level as ``the system should avoid hazardous areas.''
Given an embedding \( z_g \in \embeddings{} \) representing the goal location and a finite set of embeddings \( Z_{avoid} = \{ z_{a_1}, ..., z_{a_m} \} \) $\subseteq \embeddings{}$ (where $m = | Z_{avoid} |$), representing the hazardous areas, this requirement can be specified in ETL as:
$ \eventually(dist(z, z_g) \leq  \epsilon_g) \land_{i=1}^m \always((dist(z,z_{a_i}) > \epsilon_{a_i})  $
where the robot's location embedding is $z$, $\epsilon_g$ the minimum closeness required for reaching the goal, and  ($\epsilon_{a_1}, ...\epsilon_{a_m}$)  the minimum safe distances that must be maintained from the hazardous areas.

\paragraph{Example 2: Sequential task.} 
Another common type of tasks for robotic systems is to visit multiple goals in a sequence. This pattern is sometimes called \emph{sequenced visit} \cite{DBLP:journals/corr/abs-1901-02077}.
Given two embeddings $z_{g_1}, z_{g_2} \in \embeddings{}$ that represents the two goal locations, the requirement can be specified as an ETL specification that requires the distance from the robot and the locations to fall below a given threshold $\epsilon_g$, one after another: $ \eventually((dist(z, z_{g_1}) \leq  \epsilon_{g_1}) \land \eventually(dist(z, z_{g_2}) \leq \epsilon_{g_2}))$.

\paragraph{Example 3: Stability.}
In \textit{visual servoing} \cite{hutchinson1996tutorial}, a robot adjusts its movements to align its camera view with a target reference image, using visual feedback rather than explicit position measurements. A key requirement in such systems is \textit{stability}, ensuring that once the desired visual state is reached, the system constantly remains in the same region of the embedding space; e.g., ``once a drone reaches a desired viewpoint, it should maintain the viewpoint."
Given an embedding \( z_g \in \embeddings{} \) representing the target visual state, stability can be specified in ETL as: $  \eventually\always(dist(z, z_g) \leq \epsilon_g)$ where: \( \epsilon_g \) ensures that once the system achieves the desired visual embedding, it maintains that state indefinitely.

\subsection{Quantitative Satisfaction} 

Certain extensions of LTL, such as \emph{signal temporal logic} (STL)~\cite{Maler2004MonitoringTP}, allow for a quantitative notion of satisfaction (called \emph{robustness}~\cite{robustness}) that represents the degree to which the system satisfies or violates a specification. Among its use cases, robustness can be leveraged in \emph{optimization-based tasks}, such as trajectory planning~\cite{MPC-STL}, where the quality of candidate plans is evaluated based on how well the resulting system behavior satisfies a property (i.e., the robustness score). To enable ETL to be used for such tasks (later discussed in Section~\ref{sec:planning}), we also introduce a notion of quantitative satisfaction for ETL.

The \emph{satisfaction score} of
a trace with respect to ETL formula $\varphi$ at index $i$, bounded by $\bound{} \geq i$, is denoted
$\rob(\varphi, \trace, i, \bound)$ and defined as follows:
\begin{eqnarray*}
  \rob(u, \trace, i, \bound{}) & = & f_u(z_i) \\
  \rob(\neg \varphi, \trace, i, \bound{}) & = & - \rob(\varphi, \trace, t, \bound )\\
  \rob(\varphi_1 \land \varphi_2, \trace, i, \bound{}) & = &
                                                   \textbf{min} (\rob(\varphi_1,
                                                   \trace, i, \bound), \rob(\varphi_2,
                                                   \trace, i, \bound)
                                                   ) \\
 \rob(\always{} \varphi, \trace, i, \bound{}) & = & \textbf{inf}_{k \in [i, \bound{}]}\ \rob(\varphi, \trace, k, \bound) \\
  \rob(\eventually{} \varphi, \trace, i, \bound) & = & \textbf{sup}_{k \in [i, \bound{}]}\ \rob(\varphi, \trace, k, \bound) 
\end{eqnarray*}
where \textbf{inf} and \textbf{sup} are the infimum and supremum operators, respectively. Note that the computation of the satisfaction score is restricted to a subsequence of $\sigma$ between $i$ to $\bound{}$ (i.e., $z_i, z_{i+1},...., z_{\bound{}})$. The bound \bound{} may be determined by, for example, the planning horizon used by a planner (i.e., the length of action sequence used by the planner for behavioral prediction). For brevity, the definition for the until (\until{}) operator is omitted, but similar to the one described in~\cite{robustness}. 

\paragraph{Example.} Consider an ETL specification that describes the task of reaching a desired visual state, $\varphi \equiv \eventually(dist(z, z_g) \leq \epsilon_g)$, which is equivalent to $\eventually(\epsilon_g - dist(z, z_g) > 0 )$; i.e., $f_u(z_i) = \epsilon_g - dist(z_i, z_g)$. Suppose that the system generates a sequence of embeddings of length 4 ($\trace{} = z_0, z_1, z_2, z_3$), resulting in the following vector for $f_u$: 
$$[-0.0461393,  -0.05276561, 0.08344626, 0.0541718]$$ 
The satisfaction score of the trace, $\rob(\varphi, \trace, 0, 4)$, is the maximum value of $f_u$ between the index 0 and 4; i.e.,  0.08344626. Intuitively, the satisfaction score of $\trace$ corresponds to the point  at which the system satisfies the specification most well (i.e., it comes closest to the goal embedding).  



\subsection{Specifying Target Embeddings}

As mentioned above, ETL is specified using predicates over \emph{target embeddings} that describe parts of the physical world to be reached or avoided. In practice, the specifier (e.g., a system engineer) would not directly specify the mathematical representations of  embeddings.
Instead, a tool that uses ETL (like the planner that we introduce in Section~\ref{sec:planning}) would provide an interface through which the specifier can describe an intended ETL specification with images and/or textual descriptions for the target embeddings. The tool would then use an encoder to translate the user input into the corresponding embeddings, creating the final ETL specification.

We envision that different types of interfaces and patterns can be developed to allow the specifier to build complex ETL specifications, possibly by building on prior works in computer vision and machine learning (ML) for visual and multi-modal task descriptions.  For example, in robotics, it has been shown that categorical targets, sets of RGB images (with and without depth), and natural language descriptions can be effectively used as task descriptions~\cite{chang2023goatthing}.

\begin{figure*}[!t] 
\floatconts
  {fig:distances}
  {\caption{\textbf{Image-to-image embedding distances} We compute distance metrics between the embeddings of images captured within a Habitat scene. Each image is processed by a pretrained encoder, from which we extract the corresponding embeddings. We then calculate the pairwise distances between these embeddings using different distance metrics.}}
  {%
    \subfigure[Embedding distance metric computation for Dino features ]{%
      \label{fig:dinopic1}
      \includegraphics[width=0.7\textwidth]{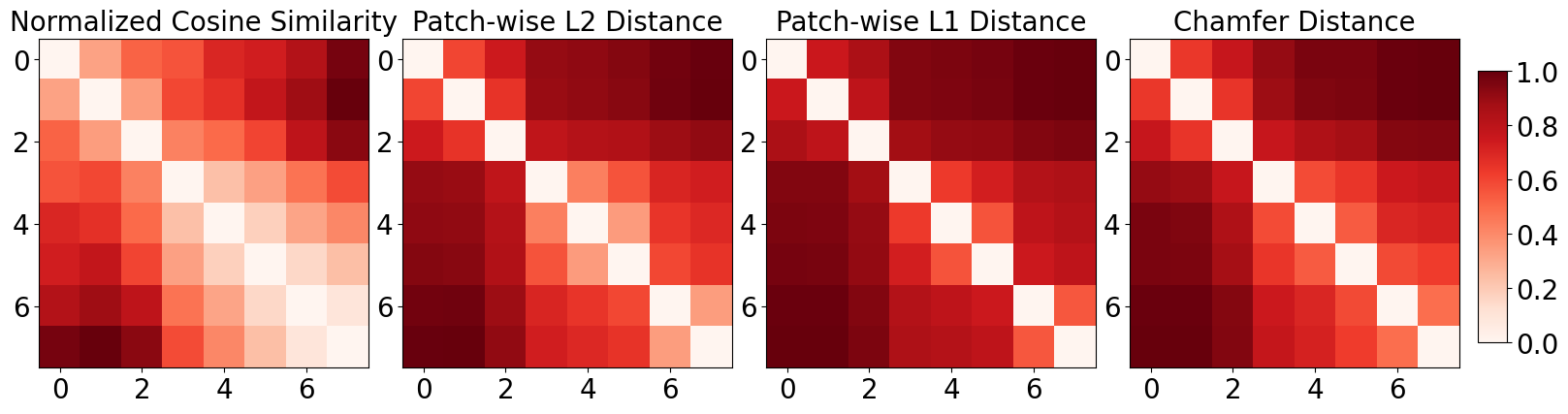}
    }\qquad 
    \subfigure[Ground truth images used for distance computation]{%
      \label{fig:dinopic2}
      \includegraphics[width=0.9\textwidth]{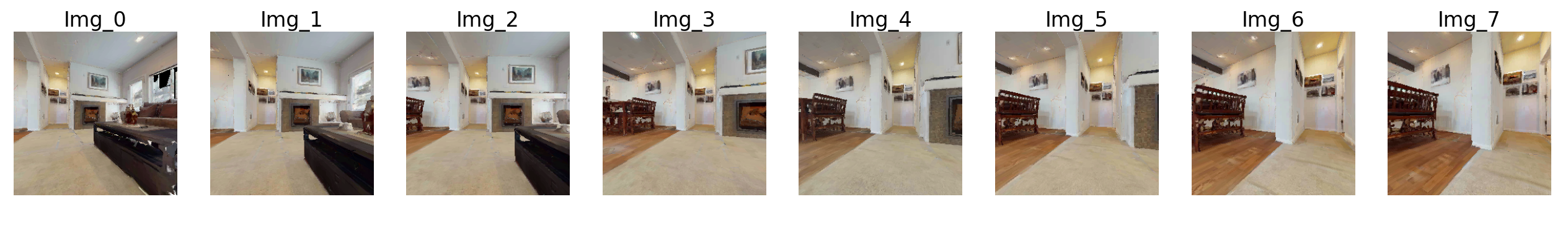}
    }
\vspace{-8mm}
  }
\end{figure*}

\subsection{Distance Functions}

\paragraph{Distance Function between Image Embeddings.}

Another decision to be made by the specifier is the choice of the distance function between a pair of embeddings. 
We have investigated four existing   distance metrics used in computer vision as  candidates for ETL: Cosine similiarity, L1, L2, and chamfer distances. Detailed descriptions of these metrics can be found in Appendix~\ref{sec:dist_im_a}.

We provide an analysis of the suitability of the metrics for capturing distinct visual states. Figure \ref{fig:distances}(a) shows the distance heatmaps for the four  metrics. For each heatmap, we compute pairwise distances between the embeddings of 8 images captured within a Habitat scene \cite{habitat19iccv} that are generated by a pretrained encoder (DinoV2), as shown in Figure \ref{fig:distances}(b).

As expected, the diagonal for all embeddings shows zero distance while off-diagonal regions form distinct block-like structures. Frames 0-2 (focused on the fireplace and seating area) cluster together while Frames 6–7 (focused on the dining area) form another cluster. Images within each cluster remain lighter in the heatmap, whereas comparing across the two clusters yields darker blocks.\footnote{A longer version of this figure with more images is available in Appendix \ref{sec:dist_im_a}.}

Additionally, the four metrics differ on their emphasis on inter‐frame distances. Cosine similarity focuses on embedding orientation, while L2, L1 and chamfer distances measure patch‐wise embedding differences. Overall, these heatmaps illustrate how even small viewpoint shifts (e.g.\ frames 2 vs.\ 3)  can be captured by an increase in embedding distance and can be used to effectively guide behavior generation. 
Note that distances over raw patch embeddings measure perceptual similarity but lack temporal context as the pretrained vision models are trained on static images. World models transform these embeddings into a learned latent space that evolves based on past states and actions. These world models are trained with a reconstruction loss and hence preserve the structure of the embedding space, ensuring that temporal relationships between embeddings transfer to the latent space. Thus, when using ETL with a world model, it is crucial to select a distance metric that aligns with the training objective of the  model.

\paragraph{Image and Textual Embeddings.}

In addition to images, target embeddings can also be specified using textual descriptions. For brevity, we refer the reader to Appendix \ref{app:text_embed} for a discussion of text-to-image embedding comparisons and their role in ETL.

\section{ETL-based Planning}
\label{sec:planning}

\begin{figure*}[t]
     \centering
    \includegraphics[width=.8\textwidth]{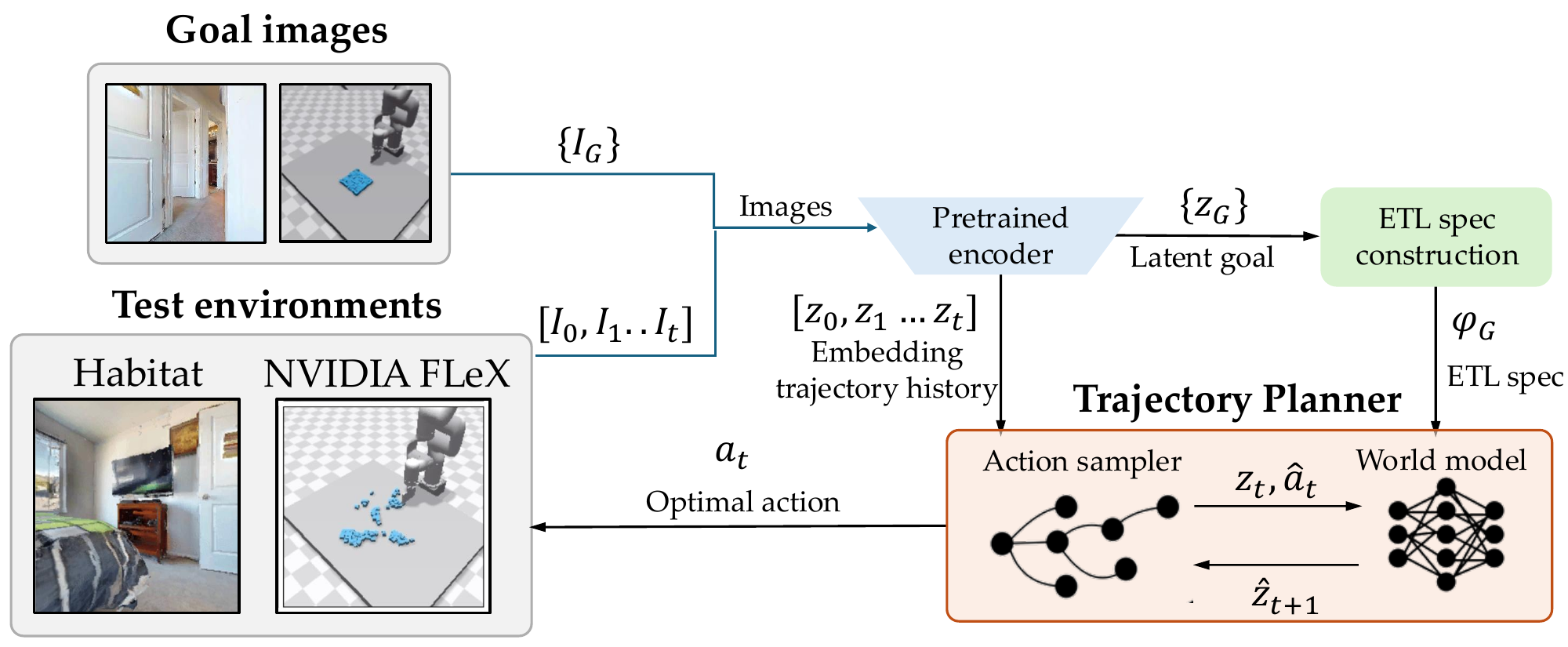}    
      \caption{
     \textbf{Overview of Planning with ETL specifications} We utilize Habitat for navigation and NVIDIA FLeX for granular manipulation. Pretrained encoders generate embeddings from goal and current observations, which are used to evaluate specification satisfaction. The planner integrates these embeddings with a world model to generate ETL satisfying actions.}
      \label{fig:overview}
\end{figure*}

In this section, we propose a method for online robotic planning that generates actions toward satisfying a given ETL specification. The high-level overview of the planning method is highlighted in Figure~\ref{fig:overview}. A key idea is for the planner to leverage a pretrained \textit{world model} when simulating and evaluating candidate actions, and then replan at each time step to perform receding horizon control. Our current prototype accepts image-based ETL specifications but a similar method can be applied for textual specifications as well.

First, our system accepts a set of \emph{goal images}, $\{I_G\}$, from the user and passes them through the pre-trained encoders (e.g., DinoV2 \cite{oquab2024dinov2learningrobustvisual}/ViT \cite{dosovitskiy2021imageworth16x16words}) to generate patch features. Concretely, an input image is divided into fixed-size patches (e.g. 16×16 pixels) that are transformed into a feature vector. These patch features capture localized visual information and are then aggregated by the encoder model to form a global image representation that will serve as reach ($z_g$) and avoid ($Z_{avoid}$) embeddings. These embeddings are used to define an ETL specification $\varphi_g$ (for examples, see Table~\ref{tab:env_table}).
During the online planning stage, we take the past and current system observations, $[I_0,\ldots, I_t]$, and pass them through the same pretrained encoders to compute a corresponding sequence of embeddings, $[z_0,\ldots, z_t]$. 

Next, the planner generates a set of candidate action sequences, $\hat{a}_t,\ldots, \hat{a}_{t+K}$ (where $K$ is the prediction horizon). Each action sequence is executed through the world model to generate a predictive trace ($\sigma$). Each predictive trace is then evaluated to compute the satisfaction score for the given specification $\varphi_g$, and the action sequence that results in the predictive trace with the highest is selected for the final plan. More formally, this planning process can be expressed as the following optimization problem: 

\begin{align}
    a_{t:t+K} &= \argmin_{\hat{a}_{t:t+K}\in \mathcal{A}_{t:t+K}} J_\varphi(z_{0:{t+K}}) 
    &\text{where} \: \hat{z}_{k+1} = p_\mathrm{\theta}(\hat{z}_k, \hat{a}_k), \quad k=t,\ldots,K-1,
\end{align}
where $J_\varphi$ is a cost function based on the desired ETL specification. For example, $J_\varphi(z_{0:{t+K}}) = \max(0, -\rho(\varphi, z_{0:{t+K}}))$. This cost function penalizes only negative satisfaction scores, i.e., traces that violate  $\varphi$ and ensures that traces satisfying $\varphi$ (where $\rho(\varphi, z_{0:{t+K}}) > 0$) have zero cost.

 In practice, the number of candidate action sequences is too large to enumerate, so our planner randomly samples $N$ action sequences\footnote{Future work can include learning a sampling distribution to increase the quality of action sequence samples.} and then select the most satisfactory one---an easily parallelizable operation. 
With the chosen action sequence, the planner applies the first action $a_t$ on the robot, and then repeats the planning process with an updated state (i.e., new sensor image). Note that the world model used here shares the same embedding space as the embeddings used in ETL; operating with a learned world model enables the direct evaluation of ETL quantitative semantics in the embedding space rather than expensively decoding back to the physical state space.

\section{Preliminary Evaluation}
\label{sec:evaluation}

We describe a preliminary evaluation to demonstrate the potential utility of ETL. In particular, we present experiments involving two types of planning tasks: (1) indoor navigation and (2) robot arm manipulation, to (1) evaluate whether embedding-based specifications can be used to generate plans that result in desirable system behaviors and (2) analyze the impact of different distance metrics.

\begin{table}[!t]
\centering
\begin{tabular}{p{0.15\linewidth} p{0.25\linewidth} p{0.45\linewidth}}
\toprule
\textbf{Environment} & \textbf{NL Description} & \textbf{ETL Specification} \\
\midrule
Navigation
& 
Visit room
& 
$ \varphi_1 = \eventually(dist(z, z_{g_1}) - \epsilon_{g_1} < 0)$
\\

Navigation
& 
Visit either room
& 
$\varphi_2 = \varphi_1 \lor  \eventually(dist(z, z_{g_2}) - \epsilon_{g_2} < 0)$ 
\\
Navigation
& 
Visit rooms sequentially
& 
$\varphi_3 = \eventually((dist(z, z_{g_2}) - \epsilon_{g_2} < 0) \land \varphi_1)$ 
\\
Manipulation
& 
Reach configuration
& 
$\psi_1 = \eventually(dist(z,z_g) - \epsilon_g <0)$
\\
Manipulation
&
Avoid 8 configurations
&
$\psi_2 = \land_{i=1}^8  \always (dist(z,z_{a_i}) + \epsilon_{a_i} > 0)$
\\
Manipulation
&
Reach and avoid
&
$\psi_3 = \psi_1 \land \psi_2$
\\
\bottomrule
\end{tabular}
\caption{ETL specifications for navigation and manipulation tasks.}
\label{tab:env_table}
\vspace{-5mm}
\end{table}

\subsection{Experimental Setup} 

For indoor navigation, we pre-train a world model on 10,000 expert traces for the PointNav task in the Habitat Environment \cite{habitat19iccv}; Habitat is a photorealistic simulator that creates indoor scenes using real-world assests.
We utilize MP3D \cite{Matterport3D} assets and build on the Perception Action Causal Transformer (PACT) from a ResNet to a ViT for it's improved preresentational capacity of ViT \cite{dosovitskiy2021imageworth16x16words}.
We focus on a sampling based planner with multiple cost functions, and plan using sequential specifications, highlighted in  Table \ref{tab:env_table}.

For manipulation, we pre-train a DinoWM world model on 2,000 traces leveraging the setup proposed in \cite{zhou2025dinowm} where granules on a flat plane are manipulated into a target goal.
To plan the actions, we utilize the integrated Dino World Model planner 
and measure success by calculating chamfer distances between the final observed image and ground truth goal image, 
similar to the DinoWM work.
We examine three cases of specifications, formalized in Table \ref{tab:env_table}.
We baseline against $\psi_1$, with a single target specification to reach; for $\psi_2$, we provide a set of targets to avoid, and in $\psi_3$, we conjunct the requirements of $\psi_1$ and $\psi_2$.

 \begin{table*}
\centering
\scalebox{1}{
\begin{tabular}{c|CCC|CCC}
\toprule
\textbf{Distance Metric} & \multicolumn{3}{c|}{\textbf{Navigation Specifications}} & \multicolumn{3}{c}{\textbf{Manipulation Specifications}} \\
\cmidrule(lr{0.7em}){2-4} \cmidrule(lr{0.7em}){5-7}
 & \textbf{$\varphi_1$} & \textbf{$\varphi_2$} & \textbf{$\varphi_3$} & \textbf{$\psi_1$} & \textbf{$\psi_2$} & \textbf{$\psi_3$} \\
\midrule
L1  &  0.0031  &  0.0005  & -0.0336  &  1.04  &  0.93  &  0.82  \\
\cdashline{1-7}
L2  &  0.0008  &  0.0035  &  0.0001  &  0.99  &  0.99  &  0.98  \\
\cdashline{1-7}
Cosine/CD  &  0.0021  & -0.0046  & -0.0274  &  1.32  &  1.32  &  1.32  \\
\bottomrule
\end{tabular}
}

\caption{\textbf{Left}: Satisfaction scores for PACT evaluations at a prediction horizon of 15 in navigation tasks. The scores  are computed using L2 since PACT was trained using L2 loss defined over predicted and ground truth embedding space. Also, L2 loss had the most granular distance separations as highlighted in Section~\ref{sec:planning}. \textbf{Right}: Final chamfer distances (CD) for manipulation tasks, computed over ground truth with a horizon length of 10 in DinoWM.}
\label{tab:combined_results}
\end{table*}

\subsection{Results and Discussion}

\paragraph{Navigation.}
, $\varphi_2$, and $\varphi_3$) when used with different base distance metrics. As highlighted in Table \ref{tab:combined_results}, for $\varphi_1$, the planner is able to achieve positive satisfaction scores with all distance metrics since it is a simple reach specification. For $\varphi_2$, we specified the target image embeddings of two different rooms disjuncted with each other, encoding the task of visiting either room. We observe that the planner performs well with L2 and L1 metrics and ends up visiting the physically closest room to its initial location. Additionally, we observed that planning with cosine similarity metric for $\varphi_2$ achieves a negative satisfaction score ($-0.0046$), possibly because it only measures high-level directionality between embedding vectors. As highlighted in Section~\ref{sec:planning}, this leads to scenes with slight perspective change being perceived as similar and hence serve as insufficient feedback for the planner.

 In order to evaluate our proposed system's ability to plan for complex tasks, we used a complex sequential visit specification  ($\varphi_3$) that involved exploring multiple rooms.
 Due to this task complexity, planning with L1 and cosine similarity distance-based specifications achieve negative satisfaction scores ($-0.0036$ and $-0.0274$, respectively). 

 For these metrics, we observed that the planner was able to satisfy the first phase of the task but failed to achieve the second phase of visiting the other room. However, with the L2-based distance metric, the planner is able to generate actions that guide it to the second room. We provide qualitative image results in Appendix \ref{app:nav}. 

In summary, the results suggest that ETL  can be used to specify desired system behaviors, and its quantitative semantics can be leveraged by a planner to steer the system towards behaviors that satisfy a given specification.

\paragraph{Manipulation.}
We performed planning for the reach ($\psi_1$), avoid ($\psi_2$), and reach-avoid ($\psi_3$) specifications using distance metrics of the L1 norm, L2 norm, and chamfer distances (CD) of the embedded images.

In Table \ref{tab:combined_results}, we present the ground truth chamfer distances computed after planning with a horizon of length 10.
We can observe that for L2 and CD, the ending distances are very similar;
however, we can also note that the results for specification $\psi_2$ shows a reduction in the final chamfer distance.
Even without a goal to plan towards, DinoWM is able to infer from the avoid embeddings that the granules should not be in a scattered configuration, and arranges them into a grouping.
From this, we hypothesize that the reach-avoid specification $\psi_3$ shows strong similarity to the reach specification $\psi_1$ due to optimization done during planning. 
From these results, we can conclude ETL specifications used during planning are able to successfully perform manipulation tasks.
An additional discussion of the results  is provided in Appendix \ref{app:man_etl}.

\section{Related Work}
There is a long line of work on specifying and verifying complex behaviors in cyber-physical and robotic systems using temporal logics, such as 
LTL, STL, and Metric Temporal Logic (MTL)~\cite{koymans1990specifying}. These logics have been used for trajectory planning \cite{kress2009temporal, Sun2022MultiagentMP, leungbackpropagation}, reinforcement learning \cite{aksaray2016qlearning, alur2023policy, 10160953}, runtime monitoring \cite{bartocci_specificationbased_2018}, and adaptive control \cite{MPC-STL, doi:10.1146/annurev-control-053018-023717, Lindemann2019ControlBF, kapoor2025stlcg++}. 

The logics outlined above can struggle with systems that rely on ML for perception, where input data can have a variable number of objects in frames and evolving bounding boxes. This has led to spatial extensions of STL and MTL that allow one to specify properties with geometric interpretations \cite{haghighi2015spatel, bortolussi2014specifying}. Specialized logics such as Timed Quality Temporal Logic (TQTL) \cite{dokhanchi2018evaluating} and Spatio-Temporal Quality Logic (STQL) \cite{hekmatnejad2021formalizing, balakrishnan2021percemon} have been proposed for perception systems that permit reasoning about properties over bounding boxes used in object detection. Recently, Spatiotemporal Perception Logic (STPL) \cite{doi:10.1177/02783649231223546} was introduced that combined TQTL with spatial logic and allows quantification over objects, as well as 2D and 3D spatial reasoning. STPL allows expressing properties by specifying relations between objects over time. Most of these logics are grounded in the symbolic outputs of a perception module (object labels, confidences, track IDs). In comparison, ETL is used to specify properties directly about the embedding space, making it easier to specify tasks such as “go to an area like this image” without depending on object detection models or their object vocabulary. 
Additionally, ETL can potentially be specified using textual descriptions as well  (using CLIP text embeddings~\cite{radford2021learning}), allowing for multimodality in  specification, in line with recent advances in pre-trained models for robotics \cite{firoozi2023foundationmodelsroboticsapplications}.

\section{Open Challenges and Future Work}

 Based on our preliminary investigation, we believe that ETL---and more generally, embeddings as a specification construct---presents a promising approach to specifying a wide range of properties about  AI-enabled systems. Beside the planning use case presented in this paper, we envision that ETL could be applied to other assurance tasks, such as online monitoring, falsification, and verification. Here, we discuss some open challenges and paths forward for  making embedding-based specifications more applicable and practical.

The benefit of operating on explicit and fully observable states (e.g., position) is that the distance metrics are spatially grounded. For instance, if the euclidean distance between a position $(x,y)$ and an obstacle is, say $1$, then that can be directly interpreted as the physical distance of one meter. However, interpreting distances in the embedding space is less spatially grounded. For instance, what does an embedding distance of $0.1$ mean between two images? A low number suggests higher similarity, but \textit{how} are they similar?
Defining thresholds for embedding distances is an open challenge for ETL. As one potential direction, we envision that existing ML techniques could be leveraged for threshold definition. For example, we could potentially calibrate or perform additional fine-tuning to ensure that the embedding distances are spatially or semantically grounded. This would involve collecting labeled data with grounded quantitative information about how similar or different two embeddings are, such as from known physical distances, or human intuition.

Another challenge is to \emph{explain} the result of an ETL-based analysis (e.g., monitoring or falsification) in terms of concepts that are human understandable, rater than in the embedding space. 
One  direction that we plan to explore is to \emph{decode} embeddings of interest (e.g., a violating trace) into the observation space (e.g., images) using a world model,  similar to how it is done in \cite{zhou2025dinowm, nakamura2025generalizingsafetycollisionavoidancelatentspace}. This would allow us to, for example, present the user with an explanation of a violation as a sequence of images, improving the interpretability of system behavior.

\bibliography{references}

\newpage
\appendix
\section{Distance Metrics over Image Embeddings}
\label{sec:dist_im_a}

We provide descriptions of the four distance metrics that we have investigated for ETL so far: 
\begin{itemize}
    \item \textbf{L1 norm.} L1 norm or manhattan distance computes the absolute differences between the corresponding elements in two embedding vectors. It is defined as: $d_{\text{L1}}(A, B) = \sum_{i} |a_i - b_i|$,
    where \( A = \{ a_i \} \) and \( B = \{ b_i \} \) represent embedding vectors. The L1 norm is a robust metric that is particularly useful in scenarios where outliers need to be ignored.
    \item \textbf{L2 norm.} L2 norm is one of the most commonly used distance metrics for embeddings. It computes the straight-line distance between two vectors and is defined as: $d_{\text{L2}}(A, B) = \sqrt{\sum_{i} (a_i - b_i)^2}$. The L2 norm effectively captures granular differences between dense embeddings and is sensitive to large variations in feature values.
    \item \textbf{Chamfer Distance} 
    Chamfer Distance (CD) is a metric used to compare two sets of points by computing the sum of minimized distances between them. Given two sets of patch embeddings, \( A = \{ a_i \}_{i=1}^{N} \) and \( B = \{ b_j \}_{j=1}^{M} \), the Chamfer Distance is defined as: $d_C(A, B) = \sum_{a \in A} \min_{b \in B} \| a - b \|_2^2 + \sum_{b \in B} \min_{a \in A} \| a - b \|_2^2$.
    This metric ensures that every patch embedding in one set is matched to the closest embedding in the other set, capturing structural similarity between the two distributions. Unlike global feature-based similarity measures, it preserves local correspondences and is hence useful for comparing high-dimensional representations such as DINOv2 patch embeddings. Note that chamfer distance is a quasi-metric as it does not satisfy the triangle inequality.
    \item \textbf{Cosine Similarity} Cosine similarity measures the cosine of the angle between two embedding vectors and captures high level directional similarity. It is defined as: 
    \[
\text{cosine similarity}(A, B) = \frac{A \cdot B}{\|A\|_2 \|B\|_2}
\]
where \( A \) and \( B \) are embedding vectors, \( A \cdot B \) is their dot product, and \( \|A\|_2 \) and \( \|B\|_2 \) are the L2 norms of the vectors. Cosine similarity ranges from -1 (completely dissimilar) to 1 (completely aligned). A value of 0 indicates the orthogonality of the two vectors. It is particularly useful for comparing the orientation of embeddings and has been successfully used to compare image similarity when used for content-based image retrieval.
\end{itemize}

The distance comparison between a larger set of images is provided in the Figure \ref{fig:distances_a}.

\begin{figure*}[htpb] 
\floatconts
  {fig:distances_a}
  {\caption{\textbf{Image-to-image embedding distances} We compute distance metrics between the embeddings of images captured within a Habitat scene. Each image is processed by a pretrained encoder, from which we extract the corresponding embeddings. We then calculate the pairwise distances between these embeddings using different distance metrics.}}
  {%
    \subfigure[Embedding distance metric computation for Dino features]{%
      \label{fig:pic1}
      \includegraphics[width=0.9\textwidth]{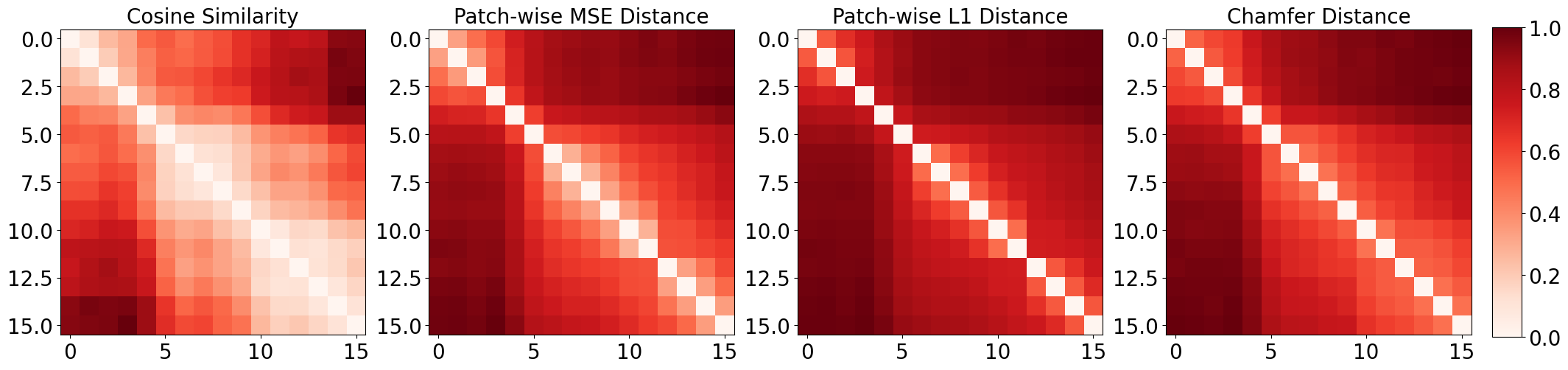}
    }\qquad 
    \subfigure[Ground truth images used for distance computation]{%
      \label{fig:pic2}
      \includegraphics[width=0.9\textwidth]{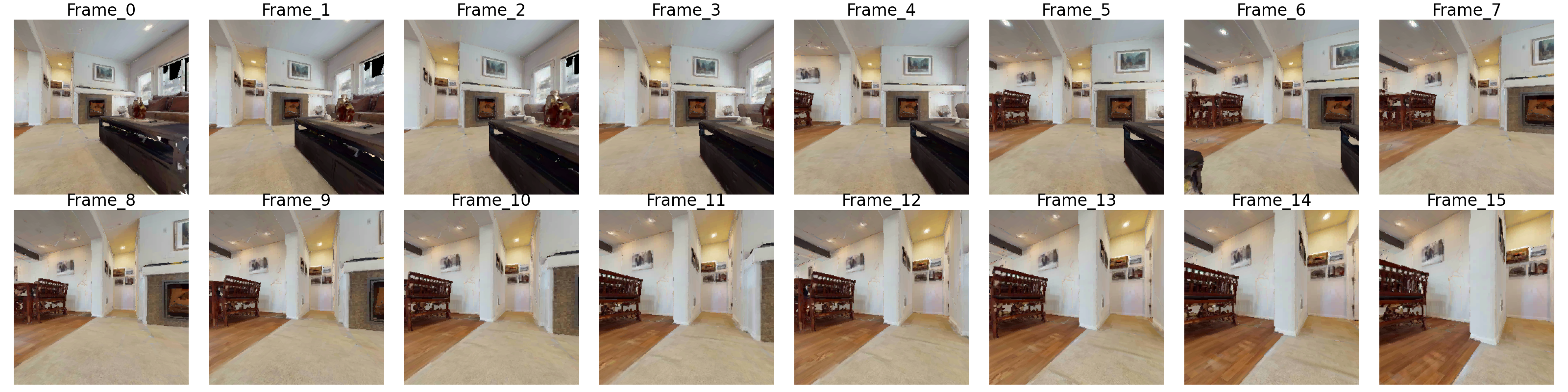}
    }
  }
\end{figure*}
\section{Textual and Image Embeddings}\label{app:text_embed}

\begin{figure*}[htbp]
     \centering
    \includegraphics[width=.8\textwidth]{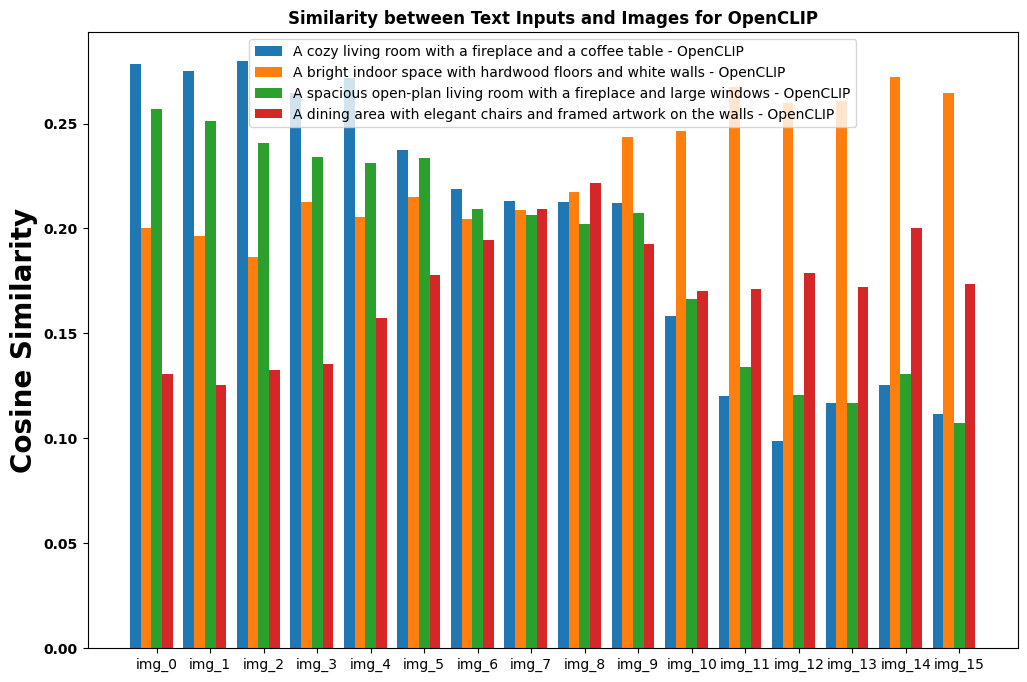}    
      \caption{
      \textbf{Text-image embedding distances} We demonstrate how models like OpenCLIP can generate textual embeddings that can be compared with image embeddings. This enables the capture of arbitrary requirements through text while serving as a reliable specification mechanism for behavior.} \label{fig:textdistances}
\end{figure*}

In this section, we talk about distance metrics between multimodal embeddings, specifically text embeddings and image embeddings. In order for a meaningful comparison between these embeddings, they both need to exist in the same latent space. Pre-trained models such as the CLIP model (Contrastive Language Image Pretraining) are trained jointly on images and texts for image captioning tasks \cite{radford2021learning}. Due to this joint pretraining, there have been multiple works that leverage CLIP's pre-trained text and image encoders for other applications due to their rich feature representation. Inspired by CLIP's original training objective that uses cosine similarities between text and image embeddings, we also define our distance metric between multimodal embeddings using cosine similarity scores. Note that cosine similarity does not satisfy the triangle inequality so it is not a metric  over the embedding space in the conventional sense but we plan to investigate other metrics that satisfy this property in future work.

We highlight our computed cosine similarity scores between different textual prompt embeddings and image embeddings in Figure \ref{fig:textdistances}. The images processed by the pretrained model are shown in Figure \ref{fig:distances_a}. We compute the text and image embeddings using OpenCLIP \cite{cherti2023reproducible} model which is a state of the art model for image captioning. We observe that the cosine similarity scores for the prompt with ``fireplace and coffee table'' is higher for images 0--7 and goes down progressively as these objects are not visible in the scene for images 8--15. Conversely, We observe increasing similarity scores for the prompt with ``hardwoord floors and white walls'' as these objects are visible in the images 8--15 and not in the previous frames. This shows that user provided textual prompts can be used directly to guide behavior generation via computing distances with image embeddings. By simply comparing prompts with scene images, an agent can prioritize actions or viewpoints that match the user's goals the most.

\section{Navigation with ETL Specifcations}
\label{app:nav}
\begin{figure*}
    \centering
    \includegraphics[width=0.7\textwidth]{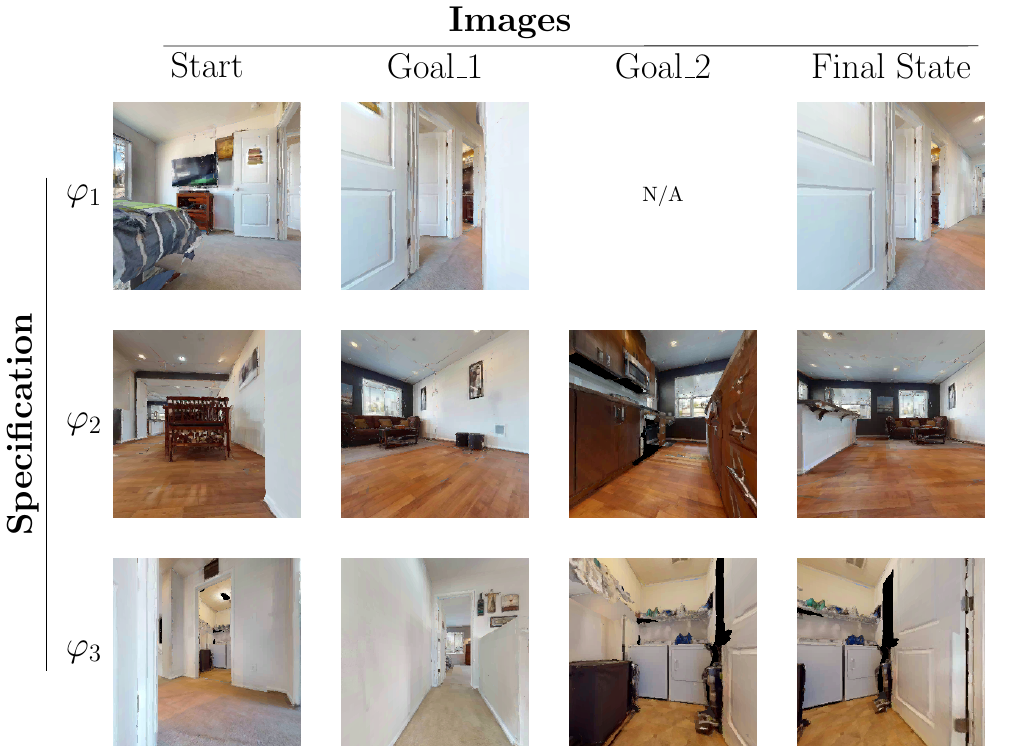}   
    \caption{Ending states after planning with a pretrained PACT world model in Habitat environment using L2 distance as the base metric.
    We present the start state as well as the goal images used for $\varphi_1$, $\varphi_2$, and $\varphi_3$ in different scenes. The planner is able to succesfully achieve positive satisfaction scores for all ETL specifications. 
    }
    \label{fig:pact_app}
\end{figure*}

We provide some qualitative results of planning with a PACT world model to satisfy ETL specifications. with L2 as the base distance metric. We picked various scenes in the habitat environment and passed target images from those environments. The start, goal and final states are highlighted in Figure \ref{fig:pact_app}. The planner is able to achieve complex behaviors like sequential visits and either-or for a variety of environments.
\section{Manipulation with ETL Specifications}\label{app:man_etl}
\begin{figure*}
    \centering
    \includegraphics[width=0.8\textwidth]{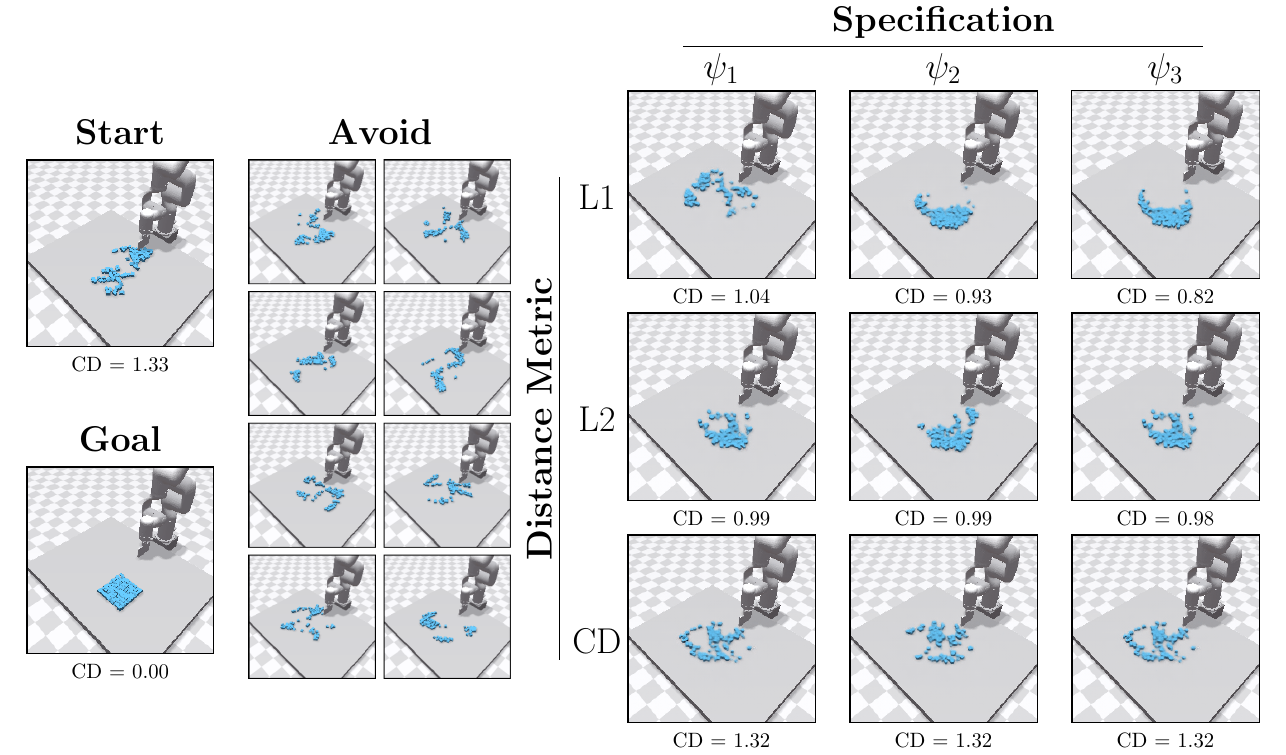}   
    \caption{Ending configurations of granules after planning with a horizon length 10 with the Dino World Model.
    We present the initial configuration as well as the target specification for $\psi_1$ and $\psi_3$, and images used for avoidance in $\psi_2$ and $\psi_3$.
    Planning was done over embedded images with L1 norm, L2 norm, and chamfer distances, with the final chamfer distance for success evaluation shown was computed against the ground truth target.
    }
    \label{fig:dino_wm_results}
\end{figure*}
We present the images used for ETL specifications in Figure \ref{fig:dino_wm_results}.
We can observe that computing chamfer distances over the embedded images does not greatly reduce the ending chamfer distance from the starting goal in the observed space, while using L1 norm and L2 norm for the distance metrics is more effective to manipulating the granules to the target specification.
Additionally, we note that there are strong similarities between the ending configurations for L2 $\psi_1$ and L2 $\psi_3$, as well as CD $\psi_1$ and CD $\psi_3$, while L1 $\psi_2$ and L1 $\psi_3$ share stronger similarities.
ETL specifications here have assisted with avoid and avoid-reach planning for manipulation.
L1 $\psi_2$ specifically shows that even without a target goal to reach, ETL specifications assist with planning towards an unknown target.
With optimization of planning for ETL specific specifications, manipulation planning can be improved from only planning with a single reach target. 
\end{document}